\documentclass[review]{elsarticle}

\usepackage{lineno,hyperref}
\usepackage{amsmath}
\usepackage{algorithm}
\usepackage{algorithmic}
\usepackage{multirow}
\usepackage{booktabs}
\usepackage{graphicx}
\usepackage{bm}
\usepackage{makecell}
\usepackage{geometry}
\journal{Journal of \LaTeX\ Templates}

\begin{document}

\begin{frontmatter}

\title{Uncertainty Guided Ensemble Self-Training for Semi-Supervised Global Field Reconstruction}

\author{Yunyang Zhang}
\author{Zhiqiang Gong}
\author{Xiaoyu Zhao}
\author{Wen Yao\corref{mycorrespondingauthor}}
\cortext[mycorrespondingauthor]{Corresponding author}
\address{Defense Innovation Institute,
	Chinese Academy of Military Science, Beijing, China\\}

\begin{abstract}
Recovering a globally accurate complex physics field from limited sensor is critical to the measurement and control in the aerospace engineering. 
General reconstruction methods for recovering the field, especially the deep learning with more parameters and better representational ability, usually require large amounts of labeled data which is unaffordable.
To solve the problem, this paper proposes Uncertainty Guided Ensemble Self-Training (UGE-ST), using plentiful unlabeled data to improve reconstruction performance.
A novel self-training framework with the ensemble teacher and pretraining student designed to improve the accuracy of the pseudo-label and remedy the impact of noise is first proposed. 
On the other hand, uncertainty-guided learning is proposed to encourage the model to focus on the highly confident regions of pseudo-labels and mitigate the effects of wrong pseudo-labeling in self-training, improving the performance of the reconstruction model.
Experiments include the pressure velocity field reconstruction of airfoil and the temperature field reconstruction of aircraft system indicate that our UGE-ST can save up to 90\% of the data with the same accuracy as supervised learning.
\end{abstract}


\end{frontmatter}


\section{Introduction}

Fast and accurate acquisition of global physics field in aerospace engineering is of great significance for stable operation of monitoring system, smooth control of the process.
Since aircraft systems are required to operate in severe environments, the direct measurement of global physics field is extremely difficult. Therefore, developing reconstruction of the global physics field based on sensor observations is essential for the measurement and control in the aerospace engineering \cite{fukami2021global, yan2016data}. Traditional reconstruction methods, including principal component regression, partial least squares (PLS) regression, support vector machine (SVM), and artiﬁcial neural network (ANN) \cite{pearson1901liii,wold1966estimation,jiang2019data,yan2004soft,desai2006soft}, is usually data-driven. However, these methods are limited by the complexity of aircraft systems, these methods are typical shallow methods with limited representational ability which can no longer be adequate for reconstructing global fields from a limited number of sensors in high-dimensional nonlinear complex physics, such as fluid dynamics \cite{kochkov2021machine}, thermodynamics \cite{hernandez2021deep}, electromagnetism \cite{puzyrev2019deep}, and solid mechanics \cite{haghighat2021physics}.

Deep learning with multiple layers has shown their potential in strong high-dimensional nonlinear problems. It can automatically learn abstract and subtle features from large amounts of data, and already demonstrated their good performance applying to physics field reconstruction \cite{fukami2021global}. 
Generally, the good performance is highly depended on plentiful labeled data. 
However, obtaining plentiful training samples for the current task is impossible.
Although massive amounts of data are accumulated during industrial processes, these data are mostly unlabeled and cannot be used directly under supervised paradigms. 
Therefore, it is significance for industrial processes to efficiently utilize these abundant unlabeled data to improve reconstruction performance.

The semi-supervised learning (SSL) method can use only limited labeled and abundant unlabeled data to train the model, and has been successfully applied in computer vision \cite{zhang2022semi}, natural language processing \cite{lai2021semi}, and other fields. 
SSL methods can be loosely classified into consistency based SSL and pseudo-label based SSL \cite{van2020survey, berthelot2019mixmatch, zhai2019s4l}.
The former SSL is mainly based on the smoothness assumption, which is not reasonable for regression probem, and therefore cannot be applied in such a typical regression problem of global field reconstruction. On the contrary, the pseudo-label based SSL, such as self-training method \cite{zou2019confidence, yang2022st++} which employ the teacher model to label the abundant unlabeled data and provides the student model for training, is flexible and not constrained by specific assumptions.   
Since self-training is well adapted to the task at hand, this paper mainly focuses on the self-training global physics field reconstruction.

Despite the good performance of the self-training method, it still mainly suffers from overfitting and noise interference. Especially in the reconstruction problem, the noise in the pseudo-label has a serious erosion on the prediction. Therefore, the unsolved question is \textbf{how to avoid the effect of inaccurate pseudo-labeling on the model performance?} 

To address the problem, this paper firstly improves the accuracy of the pseudo-label so as to reduce the damage of the pseudo-label noise on the performance of the student model. Specifically, this paper using ensemble teacher to jointly guide the training of student models. The errors in pseudo-label created by a single teacher model can be mitigated by the "collective voting" of multiple teacher models, resulting in high-quality pseudo-labels.
Secondly, the student model is guided to concentrate on the ground truth or the areas without noise, thereby eliminating noise interference and improving the performance of the student model.
During the training, the uncertainty in the pseudo-label is quantified based on the ensemble teacher, which is further used as a guided information of noise in the pseudo-label. Based on uncertainty guided learning, the student model ignores the noise region in the pseudo-label during the training process, thus avoiding the propagation and accumulation of noise between the teacher and student model. Then, in order to further reduce the interference of noise to the student model, pre-training student separates the pseudo-label from the labeled data during the training. So that the noise in the pseudo-label can be "forgetten", and the student model is forced to focus on the ground-truth to obtain higher performance.

Based on the above strategy, Uncertainty Guided Ensemble Self-Training (UGE-ST) is proposed in this paper, including ensemble teacher, uncertainty guided learning and pre-training student.
The innovations of this paper are summarized as follows: 

\begin{enumerate}
	\item This paper proposes a novel self-training framework designed to improve the reconstruction accuracy of unlabeled training data with the ensemble teacher and pre-training student that reduce the damage of the pseudo-label noise on the student model.
	
	\item This paper proposes uncertainty-guided learning for self-training to improve the performance of the model, which uses the uncertainty as the guided information of the student and supervises the learning process by reducing the effects of noise pseudo label.
	
	\item Two physics field reconstruction problems verify the effectiveness of the proposed method. Experiments show that the proposed UGE-ST in this paper is able to substantially improve model performance with limited labeled data, and is in an advanced position compared to supervision and other semi-supervised methods.
	
\end{enumerate}

The layout of this paper is given as follows. In Section II, the problem definition and the self-training framework are shortly overviewed. Then, the proposed UGE-ST are illustrated in detail, including ensemble teacher, uncertainty guided learning and pre-training student. After that, the effectiveness and feasibility of proposed approach are demonstrated in two case in Section III. Finally, conclusion is made.

\section{Methods}
\subsection{Problem Definition}
Consider an two-dimensional discretized physical field $\Gamma$ described by the governing partial differential equations:
\begin{equation}
	\bm{\dot{w}}_x=f\left(\bm{w}_x,t;\bm{\theta} \right), x\in \Gamma
\end{equation}
where $\bm{w}_x$ is the state vector in point $x$ of physical field $\Gamma$ that depends on parameters $\bm{\theta}$ and time $t$; $f$ represents the nonlinear function that governs the physical field $\Gamma$.

In practice, due to the complexity of the physical system, the state $\bm{w}$ of the whole system is usually unavailable. But the state of system on limited points can be observed by the sensors. We denote $\bm{a}(t;\bm{\theta})$ as the observed state at time $t$. 
The purpose of physical field reconstruction is to reconstruct a complete state $\bm{w}(t;\bm{\theta})$ of physical field from a limited observed state $\bm{a}(t;\bm{\theta})$,
\begin{equation}
	\bm{w}(t;\bm{\theta})=F\left(\bm{a}(t;\bm{\theta})  \right)
\end{equation}
where $\bm{a}(t;\bm{\theta})=\{\bm{a}_y, y\in \Lambda \subseteq \Gamma\}$, $\Lambda$ is the set of observed points; $F$ is the required reconstruction model which is deep neural network in this paper.
It is worth mentioning that, although the physical system is time-dependent, the reconstruction model only relies on the observed state at the current step for predicting the system states.

In order to construct the deep reconstruction model, we take the limited observation state $\bm{a}(t;\bm{\theta})$ as the input of model and the complete state $\bm{w}(t;\bm{\theta})$ as the output of model.

\begin{figure}[htbp]
	\centering 
	\includegraphics[width=0.7\textwidth]{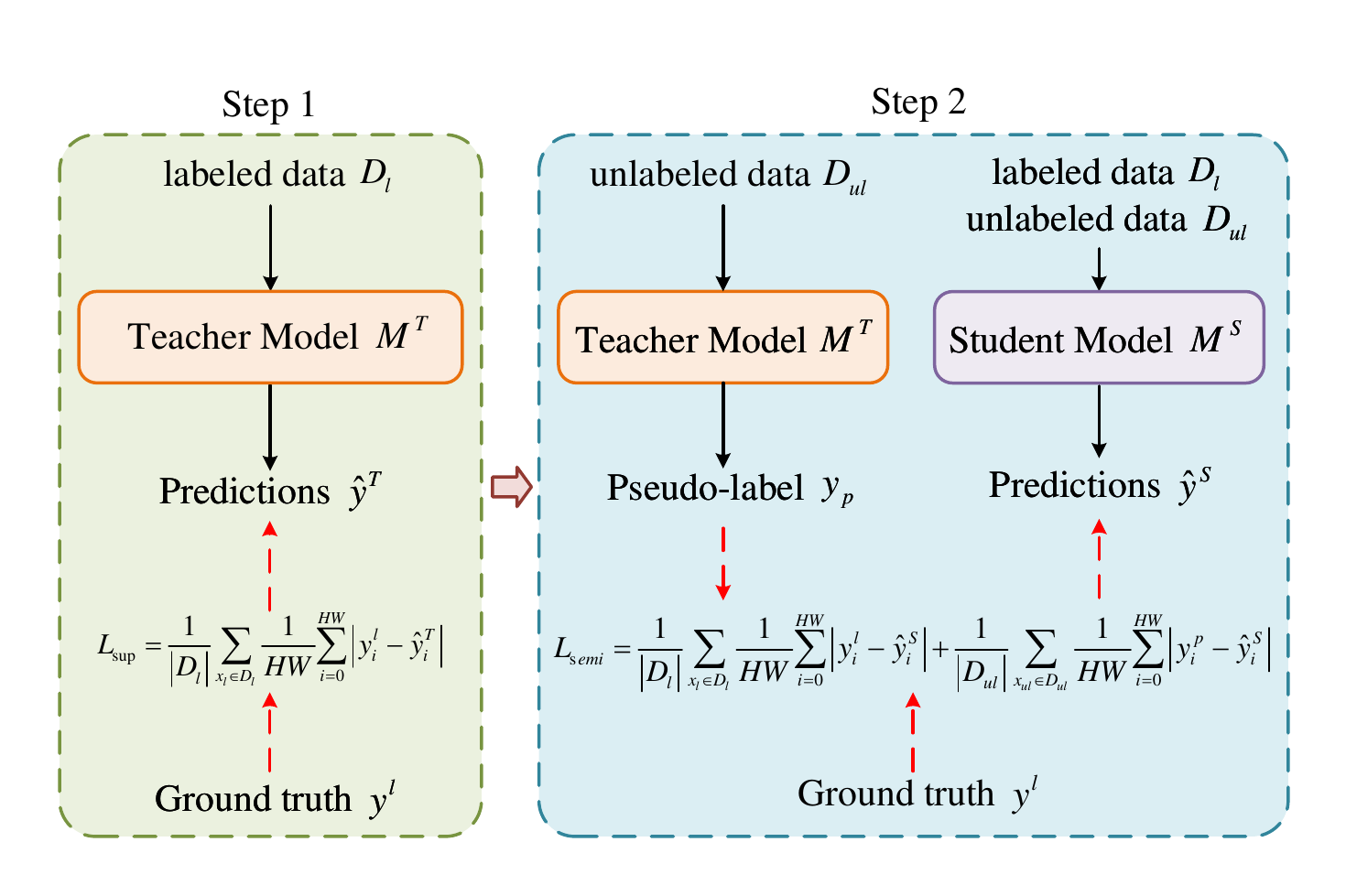} 
	\caption{Self-training method.}  
	\label{fig:1}
\end{figure}
\subsection{Self-training Framework}

Generally, the performance of the deep learning model is significantly influenced by the amount of labeled data that is manually annotated. However, collecting large amounts of data is labor-intensive and time-consuming. Semi-supervised learning is considered a promising technology to mitigate the demand for label data, thereby reducing the cost of deep learning model applications in practical engineering. 

Semi-supervised learning aims to generalize from a combination set of limited labeled data $D^l = \{(x^l_i , y^l_i)\}^N_{i=1}$ and abundant unlabeled data $D^{ul} = \{(x^{ul}_i, y^{ul}_i)\}^M_{i=1}$, where $M \gg N$. 
Self-training is a classic semi-supervised learning method based on the idea of pseudo-label. The main steps are divided into two steps, as shown in the Fig.~\ref{fig:1}:
\begin{enumerate}
	\item Firstly, a small amount of labeled data $D_l$ is used to train the teacher model $M^T$ with $L_1$ loss, which is formulated as:
	\begin{equation}
		{L}_{sup }=\frac{1}{\left| {{D}_{l}} \right|}\sum\limits_{{{x}_{l}}\in {{D}_{l}}}{\frac{1}{HW}}\sum\limits_{i=0}^{HW}{\left| y_{i}^{l}-\hat{y}_{i}^{T} \right|}
		\label{eq:3}
	\end{equation}
	where $y_{i}^{l}$ and $\hat{y}_{i}^{T}$ represent the ground truth and the prediction of teacher model, $W$ and $H$ represent the width and height of physical field.
	
	\item Secondly, training the student model $M^S$ through the teacher model $M^T$. The teacher model $M^T$ is used to predict the unlabeled data $D_{ul}$, and the predictions are adopted as the pseudo-label of unlabeled data. At the same time, labeled data is combined with pseudo-label to train the student model. The constraint can be formulated as:
	\begin{equation}
		\begin{split}
			{{L}_{{semi}}}=\frac{1}{\left| {{D}_{l}} \right|}\sum\limits_{{{x}_{l}}\in {{D}_{l}}}{\frac{1}{HW}}\sum\limits_{i=0}^{HW}{\left| y_{i}^{l}-\hat{y}_{i}^{S} \right|} \\
			+\frac{1}{\left| {{D}_{ul}} \right|}\sum\limits_{{{x}_{ul}}\in {{D}_{ul}}}{\frac{1}{HW}}\sum\limits_{i=0}^{HW}{\left| y_{i}^{p}-\hat{y}_{i}^{S} \right|}
		\end{split}
	\end{equation}
	where $\hat{y}_{i}^{S}$ is the prediction of student model.
\end{enumerate}

\emph{Discussion about self-training.} 
Unlike the consistency regularization method, the self-training method does not depend on the smoothness assumption and is simple and versatile. However, the student model is trained based on the pseudo-label predicted by the teacher model. When there is a large noise in the pseudo-label, the performance of the student model is affected. 

In order to avoid the influence of false label noise on the student model, we improve the self-training method in two aspects. The direct idea is to improve the accuracy of pseudo labels as much as possible so that the noise in pseudo labels is as small as possible to avoid noise's influence on the performance degradation of student models. Another idea is to prevent the effect of noise in pseudo labels on student model as much as possible when the accuracy of pseudo labels is determined. 

\begin{figure*}[htbp]
	\centering 
	\includegraphics[width=\textwidth]{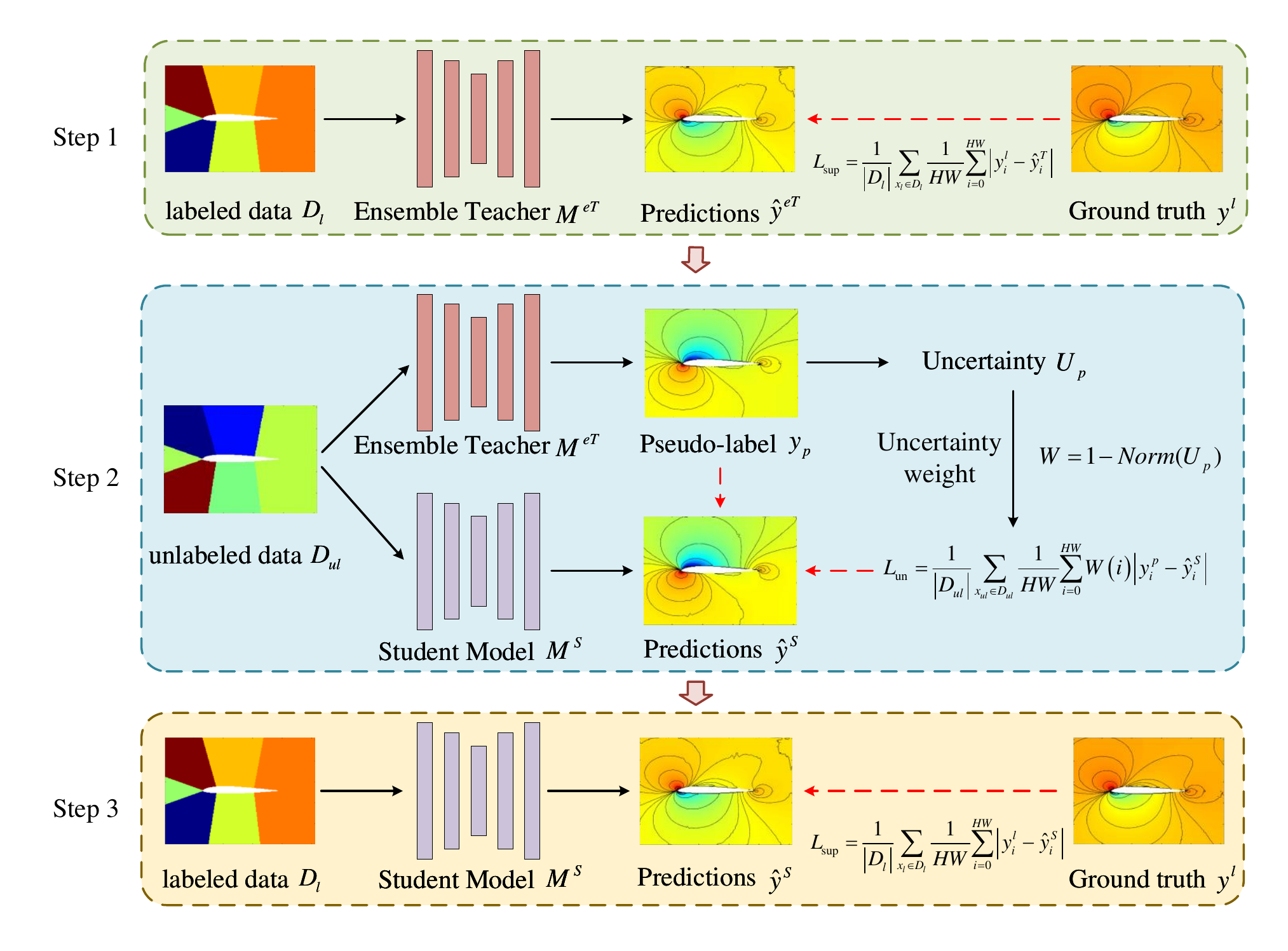} 
	\caption{Uncertainty guided ensemble self-training method.}  
	\label{fig:2}
\end{figure*}
\subsection{Uncertainty Guided Ensemble Self-Training}
This paper proposes UGE-ST including ensemble teacher, uncertainty guided learning, and pre-training student to improve the performance of reconstruction. The pre-training student and the uncertainty-guided learning improve the performance of the student model by avoiding the influence of noise in pseudo-label as much as possible. While the ensemble teachers directly improve the accuracy of pseudo-label, further enhances the performance of the student model. The proposed method framework is shown in Fig.~\ref{fig:2}. Compared with the basic self-training method, our method consists of three steps: 

\begin{enumerate}
	\item Firstly, a small amount of labeled data $D_l$ is used to train the ensemble teacher model $M^{eT}$ with $L_1$ loss, which is formulated in Eq.~\ref{eq:3}.
	
	\item Secondly, similar with basis self-training, the student model $M^S$ is trained by the ensemble teacher $M^{eT}$. The predictions of the ensemble teacher $M^{eT}$ on the unlabeled data $D_{ul}$ are adopted as the pseudo-label. Then, the uncertainty of the pseudo-label is quantified and normalized as weights to be multiplied with pseudo-label supervision constraint, getting the uncertainty-guided learning constraint ${L}_{{un}}$.
	
	\item Thirdly, a small amount of labeled data $D_l$ is employed to re-train the student model $M^S$ with $L_1$ loss.
\end{enumerate}

\subsubsection{Ensemble teacher}
Ensemble learning is a classic idea in deep learning, which combines multiple weakly supervised models in order to obtain a better and more comprehensive, strongly supervised model. The potential idea of ensemble learning is that even if a weak model gets a wrong prediction, other models can correct the error. Generally speaking, the performance of the model obtained by ensemble learning is better than that of a single model. Due to the scarcity of labeled data for teacher model training, the performance of a single teacher model is poor, resulting in large pseudo-label noise. Here we combine ensemble learning with self-training, using ensemble learning to train and combine multiple teacher models to obtain a more accurate pseudo-label. 

Specifically,  
We first initialize multiple teacher models $\left\{ M_{1}^{T},M_{2}^{T},\cdots ,M_{n}^{T} \right\}$. Multiple teacher models are trained by labeled data $D_l$ follow the supervised loss as shown in Eq.~\ref{eq:3} to obtain ensemble teachers ${{M}^{eT}}=\left\{ M_{1}^{T},M_{2}^{T},\cdots ,M_{n}^{T} \right\}$.
The unlabeled data $D_{ul}$ is predicted by ensmeble teachers ${M}^{eT}$ acquiring results $\left\{ \hat{y}_{1}^{eT},\hat{y}_{2}^{eT},\cdots ,\hat{y}_{n}^{eT} \right\}$, where $\hat{y}_{i}^{eT}={M_i}^{T}(D_{ul})$.
Then, average multiple predictions to obtain pseudo-label ${y}_{p}$ for the unlabeled data as follows:
\begin{equation}
	{{y}_{p}}=mean\left( \hat{y}_{i}^{eT} \right),i=1,2,...,n
\end{equation}

\subsubsection{Uncertainty guided learning}
Although using ensemble teachers can improve the accuracy of pseudo-label, noise still exists and affects the training of student model. We expect to filter out the noise in the pseudo-label so that the student model can learn more areas with low or no noise. Here we propose using uncertainty to guide student model training. Uncertainty can reflect the noise in the model prediction results, usually, the area with large noise is also uncertain. 
Since multiple different predictions $\left\{ \hat{y}_{1}^{eT},\hat{y}_{2}^{eT},\cdots ,\hat{y}_{n}^{eT} \right\}$ can be obtained for the same sample by ensemble teachers ${M}^{eT}$, we naturally use variance to measure the uncertainty of pseudo-label ${y}_{p}$ predicted by the teacher model as follows:
\begin{equation}
	{{U}_{p}}=var\left( \hat{y}_{i}^{eT} \right),i=1,2,...,n
\end{equation}

Then, this study normalizes the uncertainty $U_p$ into range $(0, 1)$, and weight the pixel value according to the uncertainty of each pixel in the pseudo-label. 
Among them, regions with larger uncertainty are given smaller weights, and areas with smaller uncertainty are given larger weights. The large weights force the model to learn more areas with smaller noise, while the smaller weights ignore regions with larger noise, thus avoiding the influence of noise on the model during the learning process. This paper defines the uncertainty weights as follows:
\begin{equation}
	W=1-Norm(U_p)
\end{equation}

In the end, combining uncertainty weights and pseudo-label loss to obtain uncertainty guided learning loss: 
\begin{equation}
	{{L}_{{un}}}=
	\frac{1}{\left| {{D}_{ul}} \right|}\sum\limits_{{{x}_{ul}}\in {{D}_{ul}}}{\frac{1}{HW}}\sum\limits_{i=0}^{HW}{W(i)\left| y_{i}^{p}-\hat{y}_{i}^{S} \right|}
	\label{eq:8}
\end{equation}

\subsubsection{Pre-training student}
The training of neural networks relies on empirical risk minimization. The more data used for training, the more consistent the distribution of datasets with that of all data, and the closer the empirical risk is to the expected risk.
When the amount of data is small, the empirical risk effect is not ideal because empirical risk minimization tends to bring about overfitting. Therefore, it is necessary to expand the dataset so that the training data distribution is as consistent as possible with the full data distribution. Although the self-training introduces a large amount of pseudo-label data, the presence of noise in the pseudo-label mistake the distribution with that of the real data, which eventually causes the empirical risk minimization to fail.

In order to avoid the interference of pseudo-label, this paper proposes a two-stage approach to train the student model, namely pre-training student. The idea is to exploit the catastrophic forgetting phenomenon in neural networks. 
First, the student model $M^S$ is pre-trained using pseudo-label $y_p$, and the loss function is shown as Eq.~\ref{eq:8}.
The student model $M^S$ is then retrained using a small amount of labeled data $D_l$, the constraint can be formulated as:
\begin{equation}
	{{L}_{{sup}}}=
	\frac{1}{\left| {{D}_{l}} \right|}\sum\limits_{{{x}_{l}}\in {{D}_{l}}}{\frac{1}{HW}}\sum\limits_{i=0}^{HW}{\left| y_{i}^{l}-\hat{y}_{i}^{S} \right|}
	\label{eq:8}
\end{equation}

\section{Experiments}
In this section, the effectiveness of the proposed STP method was verified through its application to two study cases. The first case is the airfoil velocity and pressure field reconstruction, and the second is the electronic devices temperature field reconstruction. In order to demonstrate the superiority of the proposed UGE-ST, the fully and semi supervised learning methods based on deep learning is implemented to compare performance, where the semi supervised methods include Mean teacher \cite{tarvainen2017mean}, co-training\cite{blum1998combining}, and the vanilla self-training \cite{yang2022st++}, and the rest remains unified. 

\subsection{Airfoil velocity and pressure field reconstruction}
\subsubsection{Background and experimental setting}
Reconstructing the pressure and velocity field of airfoil based on the finite sensors is significant to the design of the airfoil. This section adopts airfoil data \cite{thuerey2020deep} to verify the validity of proposed method.
The convolutional neural network (CNN) is employed to implement the proposed UGE-ST method. CNN is the popularly deep learning model, which is widely used in computer vision \cite{gong2020statistical, gong2019cnn}. Compared with Multilayer Perceptron (MLP), CNN with smaller parameters has the ability to process spatial information. Thus it is suitable for processing regular physical field data. Here we adopt U-net \cite{ronneberger2015u} as the backbone of model. U-net is an effective CNN structure for image-to-image regression. It can capture the overall and detailed features of the image, and has the advantages of multi-scale fusion and processing large images. 

\begin{figure}[htbp]
	\centering 
	\includegraphics[width=0.6\textwidth]{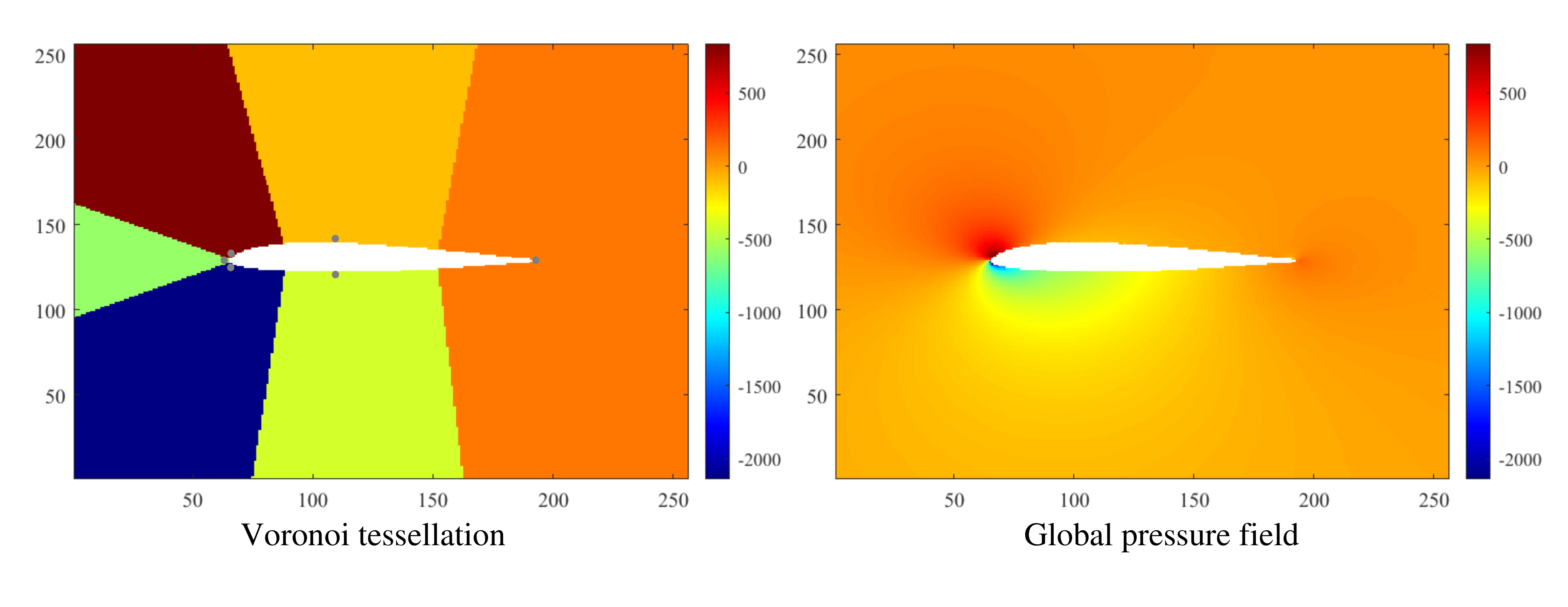} 
	\caption{Voronoi tessellation and global pressure field.}  
	\label{fig:3}
\end{figure}
To use a CNN framework, the sparse observation data needs to be projected into an image in an appropriate manner.
Similar to \cite{fukami2021global}, this paper maps local measurements to the spatial domain via Voronoi tessellation \cite{voronoi1908new} as shown in the left of Fig.~\ref{fig:3}. The grey dots in the figure indicate the placed sensors.
Compared with the form of sparse measurement as input directly, voronoi tessellation can retain the spatial information of the sensor measurement points. Besides, the output of deep learning model is the global velocity or pressure field of airfoil as shown in the right of Fig.~\ref{fig:3}. 

In this case, a total of 1200 labeled data and 800 unlabeled data are generated using finite element simulation. In order to fully verify the applicability of the proposed method, we divides the labeled data into different scale partition protocols include 25, 50, 100, 200, 400, and 800. In addition, 400 labeled samples are set aside as test data. 

In order to evaluate the performance of methods, we employ the Mean Absolute Error (MAE) as the evaluation metric, which is expressed as:

\begin{equation}
	\text{MAE}=\frac{1}{N}\sum_{i=1}^N\left\vert \hat{y}_{i}-y_{i} \right\vert
\end{equation}
where $y_{i}$ is the ground truth, $\hat{y}$ is the prediction, $N$ is the number of samples.

All experiments are implemented based on the Pytorch framework. The model training is completed on a high-performance computer server, and its computing resource allocation is Intel(R) Xeon(R) Gold 6242 CPU @ 2.80 GHz, Nvidia GTX 3090 GPU with 24GB vRAM, and 500 GB RAM. We initialize the weights of whole network randomly and train the models with AdamW optimizer. In order to ensure the fairness, the parameters of the optimizer used in experiments are consistent. The initial learning rate is $\eta = 0.001$. Besides, Cosine Annealing Warm Restarts scheduler is selected as our learning rate policy. 
In all experiments, the epoch is fixed as 100, the batch size is set to 8 for both labeled and unlabeled samples.
\subsubsection{Prediction Results and Analysis} 

\setlength{\tabcolsep}{3pt}
\begin{table}[htbp]
	\centering
	\caption{Performance comparison under different number of labeled data for the airfoils velocity and pressure field.}
	\begin{tabular}{cc|cccccc}
		\toprule
		\multicolumn{2}{c|}{Number of labeled data} & 25    & 50    & 100   & 200   & 400 &800\\
		\midrule
		\multirow{3}{*}{Supervision} & Pressure & 2.350 & 0.745 & 0.599 & 0.442 & 0.402 & 0.376\\
		& $x$-axis velocity & 0.543 & 0.170 & 0.105 & 0.087 & 0.072 & 0.065\\
		& $y$-axis velocity & 0.128 & 0.508 & 0.034 & 0.024 & 0.021 & 0.020\\
		\midrule
		\multirow{3}{*}{Mean teacher} & Pressure& 14.71 & 6.510 & 2.570 & 1.091 & 0.853 & 0.890 \\
		& $x$-axis velocity & 2.049 & 0.684 & 0.297 & 0.155 & 0.139 & 0.127\\
		& $y$-axis velocity & 0.842 & 0.391 & 0.125 & 0.059 & 0.056 & 0.042\\
		\midrule
		\multirow{3}{*}{Co-training} & Pressure& 2.495 & 0.875 & 0.745 & 0.614 & 0.515 & 0.443\\
		& $x$-axis velocity & 0.549 & 0.168 & 0.118 & 0.105 & 0.093 & 0.095\\
		& $y$-axis velocity & 0.138 & 0.056 & 0.035 & 0.027 & 0.024 & 0.027\\
		\midrule
		\multirow{3}{*}{Self-training}    & Pressure& 2.298 & 0.714 & 0.531 & 0.385 & 0.347 & 0.337 \\
		& $x$-axis velocity & 0.538 & 0.161 & 0.097 & 0.079 & 0.063 & 0.059\\
		& $y$-axis velocity & \textbf{0.125} & 0.048 & 0.029 & 0.021 & 0.017 & 0.017\\
		\midrule
		\multirow{3}{*}{UGE-ST(Ours)}   & Pressure& \textbf{1.637} & \textbf{0.690} & \textbf{0.425} & \textbf{0.298} & \textbf{0.277} & \textbf{0.281} \\
		& $x$-axis velocity & \textbf{0.429} & \textbf{0.121} & \textbf{0.064} & \textbf{0.047} & \textbf{0.037} & \textbf{0.042}\\
		& $y$-axis velocity & 0.128 & \textbf{0.032} & \textbf{0.019} & \textbf{0.014} & \textbf{0.013} & \textbf{0.013}\\
		\bottomrule
	\end{tabular}%
	\label{tab:1}%
\end{table}%
The prediction results concerning MAE are tabulated in Tab.~\ref{tab:1}, and the main purpose is to compare the prediction performances of different methods.
The results shows that the proposed UGE-ST significantly exceeds the supervised baseline and other semi-supervised methods regardless of the number of labels. Taking pressure field prediction as an example, the performance of UGE-ST surpass the supervised baseline by 30\%, 29\%, 33\%, 31\%, and 25\%, under 25, 100, 200, 400, and 800 labeled data respectively. Compared with the self-training, UGE-ST acquired 29\%, 20\%, 23\%, 20\%, and 17\% improvements, respectively.
As for the velocity in $x$ and $y$ axis, our method also achieves a better performance.
Besides, the results also shows that the number of labeled data affects the model prediction accuracy, and the performance of all models decays significantly with decreasing of labels number. 

It is worth mentioning that other semi-supervised methods, such as Mean teacher and Co-training, achieve poor prediction accuracy, even worse than the supervised baseline. The reason is that such end-to-end methods receive poor performance during the training process's early stage, further spoilt pseudo-label with large noise. As the iterative learning process proceeds, the noise in the pseudo-label accumulates, disrupting the learning process of the model and eventually leading to a degradation of the model's performance. 
The self-training is a two-stage training approach, in which the teacher model is first trained with labeled data, and then the student model is guided by the teacher model. As the trained teacher model contains less noise than that in the early training stage, the performance degradation caused by error accumulation is mitigated.
In other words, although there is still noise affecting the student model, this two-stage training approach ensures the performance of student is theoretically not lower than that of the teacher model.
And our method can further improve the performance of the teacher model with the help of ensemble learning on the base of self-training, while avoiding the noise in the teacher model to affect the learning of the student model by the uncertainty guided learning and the pre-training student. 
\begin{figure}[htbp]
	\centering 
	\includegraphics[width=0.7\textwidth]{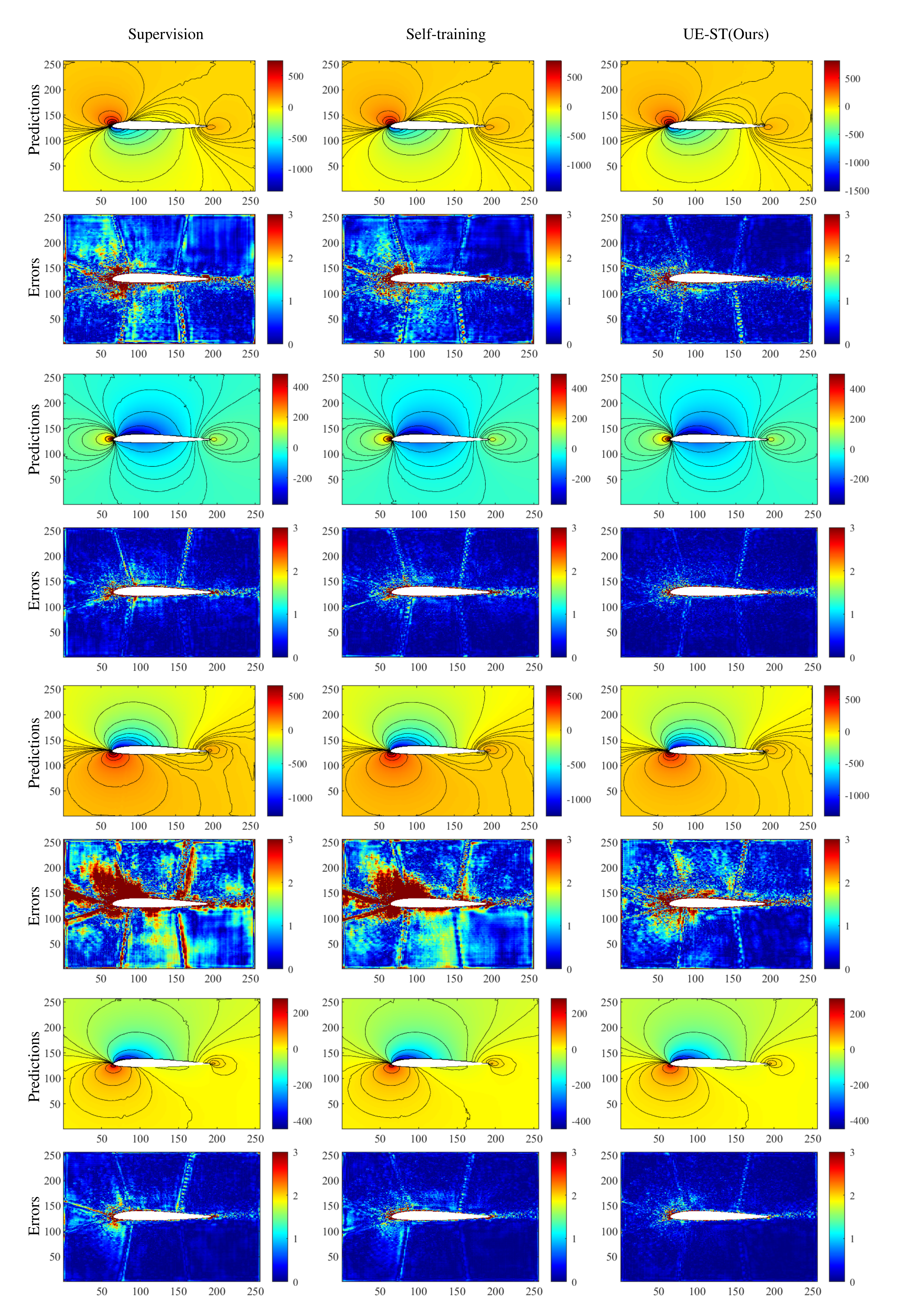} 
	\caption{Visualization of predictions and errors for pressure field.}  
	\label{fig:4}
\end{figure}

The predictions and errors visualization of supervision, self-training and UGE-ST for the pressure field are shown in Fig.~\ref{fig:4}. As seen from that, all three methods can predict the trend of the pressure field, and our approach can achieve less error. As shown in row six of Fig.~\ref{fig:4}, the predictions of the self-training and supervised methods have large errors around the airfoil, while our approach suppresses the errors well.

\subsection{Electronic devices temperature field reconstruction}
\subsubsection{Background and experimental setting}
The normal work of aircraft systems highly depends on the stable environment temperature and heat dissipation is essential to guarantee the working environment due to the internally generated heat.
Thermal management of aircraft systems is an effective way to guarantee the proper working environment. Temperature field reconstruction \cite{chen2021machine} is a base task to obtain the real-time working environment of aircraft systems, which is adopt to verify the performance of our method.

\begin{figure}[htbp]
	\centering 
	\includegraphics[width=0.6\textwidth]{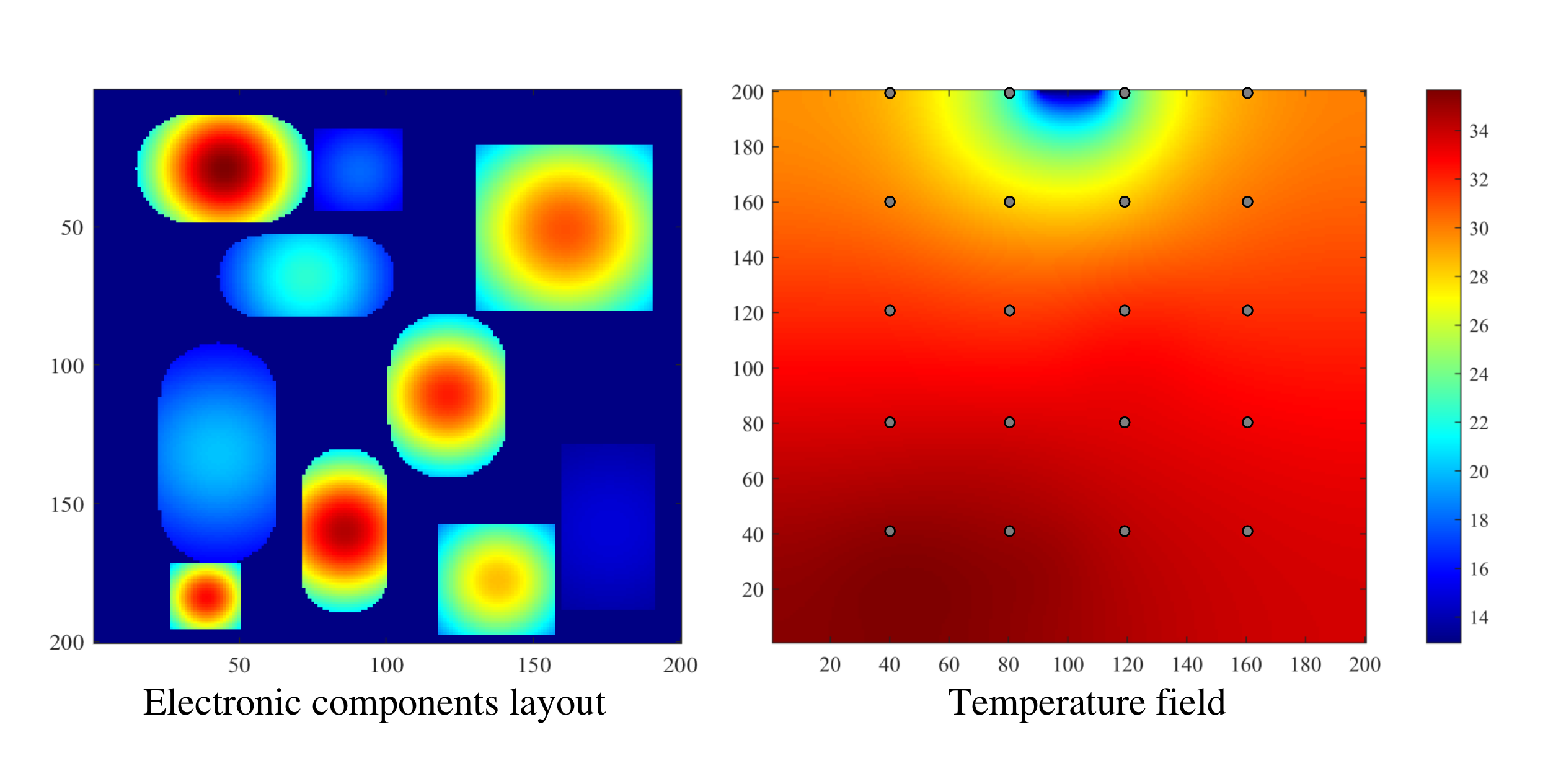} 
	\caption{Sensors location and temperature field.}  
	\label{fig:5}
\end{figure}
To verify the generality of the proposed method, in this case we employ another classical neural network, viz. MLP, to implement our UGE-ST. 
The temperature field data is shown in the Fig.~\ref{fig:5}. The grey dots in the figure indicate the placed sensors.
MLP directly accepts sparse observations of the temperature field as model inputs while outputting the temperature value of points in the whole field. The temperature field data size used in this paper is $200 * 200$, the number of placed sensors are 20, and we construct a MLP with 5 layers whose structure is shown in the Tab.~\ref{tab:2} .
\setlength{\tabcolsep}{9pt}
\begin{table*}[htbp]
	\centering
	\caption{The structure of MLP.}
	\begin{tabular}{c|ccccc}
		\toprule
		Layers & Input layer & Hidden layer 1  & Hidden layer 2 & Hidden layer 3 & Output layer \\
		\midrule
		Number of neurons & 20 & 128 & 1280 & 4800 & 40000 \\
		\bottomrule
	\end{tabular}%
	\label{tab:2}%
\end{table*}%

In this case, a total of 1500 labeled data and 2000 unlabeled data are generated using finite element simulation. Similar to the first study case, we divides the labeled data into different scale partition protocols include 25, 50, 100, 200, 500, and 1000. Besides, 500 labeled samples are set aside as test data. 

This section also uses MAE as the evaluation metric. All experimental settings are consistent with the first case, except that the epoch is set to 80.

\subsubsection{Prediction Results and Analysis}
The prediction results concerning MAE are tabulated in Tab.~\ref{tab:3}, and the main purpose is to test the generalization and superiority of our method.
The results shows that the proposed UGE-ST significantly exceeds other methods in different cases. The performance of UGE-ST surpass the supervised baseline by 33\%, 35\%, 36\%, 30\%, 25\%, and 24\%, under 25, 50, 100, 200, 500, and 1000 labeled data, respectively. Compared with the self-training, UGE-ST acquired 31\%, 29\%, 25\%, 17\%, 20\%, and 14\% improvements, respectively.
Besides, the results also shows that with the decrease of labeled data numbers, the gain obtained by our method is increasing. In another words, our method can also achieve competitive  results in the case of few shot.

From the perspective of changes in the number of labeled data, our method achieves an accuracy of 6.018e-03 with 100 samples, lower than 6.948e-03 acquired by the supervised method under 1000 samples. The observation shows that our approach can greatly reduce the amount of labeled data required for training with the same performance. 
The proposed method can save ten times the number of samples in some cases and at least two times the number of samples.

\begin{figure}[htbp]
	\centering 
	\includegraphics[width=0.7\textwidth]{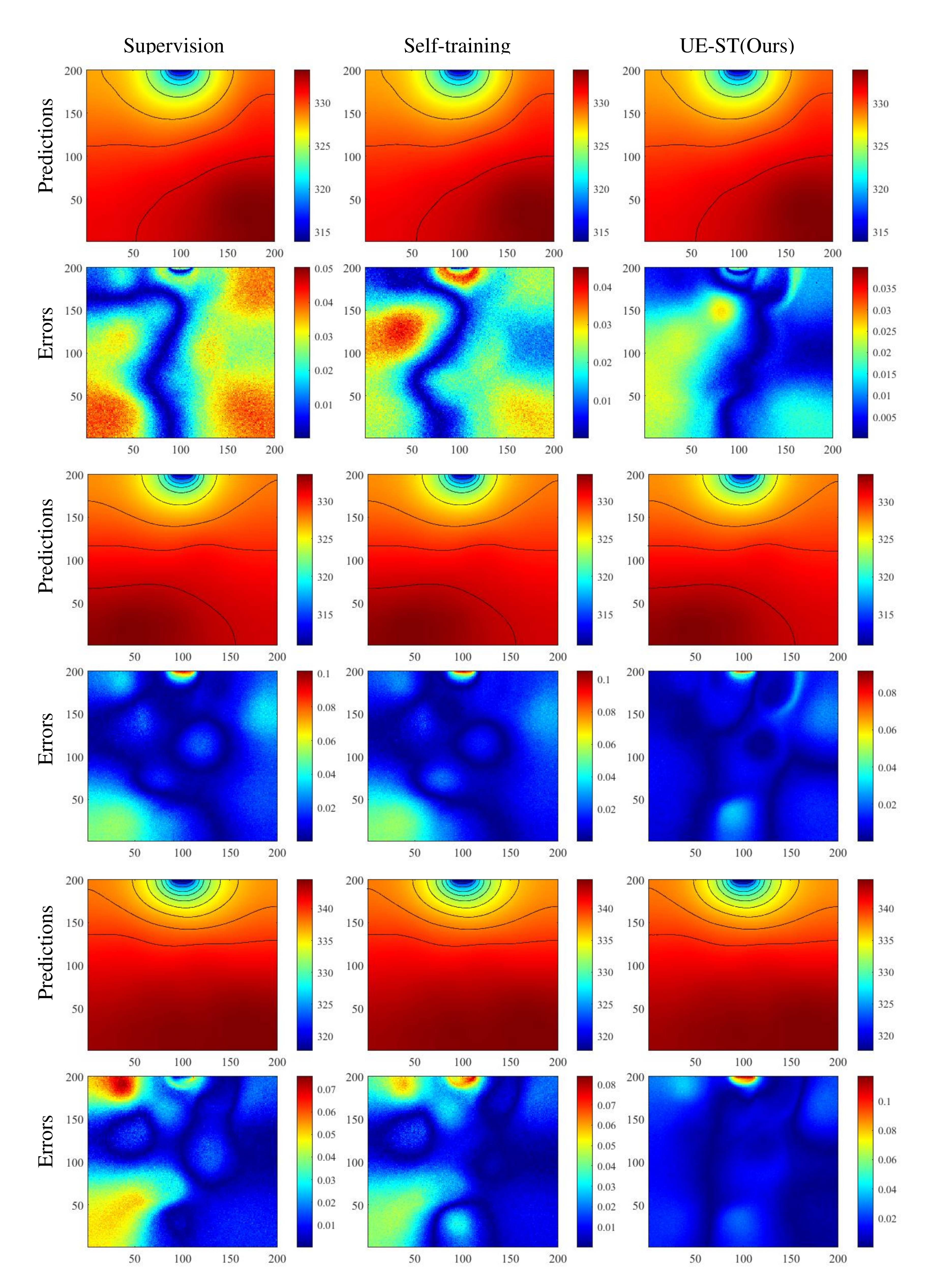} 
	\caption{Visualization of predictions and errors for temperature field}  
	\label{fig:6}
\end{figure}

The predictions and errors visualization of supervision, self-training and UGE-ST for the temperature field are shown in Fig.~\ref{fig:6}. Although supervision and self-training can predict the trend of the temperature field, large errors still exist. In comparison, the proposed method can outstandingly reduce the error in these regions. As shown in row six of Fig.~\ref{fig:6}, the predictions of the self-training and supervised methods have large errors in the lower left corner, while our approach suppresses the errors well.

\setlength{\tabcolsep}{4pt}
\begin{table*}[htbp]
	\centering
	\caption{Performance comparison under different number of labeled data for the temperature field.}
	\begin{tabular}{c|cccccc}
		\toprule
		Number of labeled data & 25    & 50    & 100   & 200   & 500 &1000\\
		\midrule
		Supervision & 1.718e-02 & 1.224e-02 & 9.403e-03 & 7.887e-03 & 7.241e-03 & 6.948e-03\\
		\noalign{\smallskip}
		Mean teacher & 2.178e-02 & 1.974e-02 & 1.315e-02 & 1.033e-02 & 9.225e-03 & 6.779e-03 \\
		\noalign{\smallskip}
		Co-training & 2.085e-02 & 1.207e-02 & 8.550e-03 & 6.717e-03 & 6.124e-03 & 6.580e-03\\
		\noalign{\smallskip}
		Self-training & 1.654e-02 & 1.120e-02  & 8.054e-03  & 6.624e-03 & 6.786e-03 & 6.087e-03 \\
		\noalign{\smallskip}
		UGE-ST(Ours)& \textbf{1.146e-02} & \textbf{7.942e-03} & \textbf{6.018e-03} & \textbf{5.529e-03} & \textbf{5.431e-03} & \textbf{5.262e-03} \\
		\bottomrule
	\end{tabular}%
	\label{tab:3}%
\end{table*}%

\subsection{Ablation Studies}
This section conducts the ablation studies to exhibit the roles of ensemble teachers, uncertainty guided learning and pre-training student. All the experiments are run based on the temperature field reconstruction task with $200$ labeled data. 
\subsubsection{The influence of ensemble teachers number}
\begin{figure}[htbp]
	\centering 
	\includegraphics[width=0.6\textwidth]{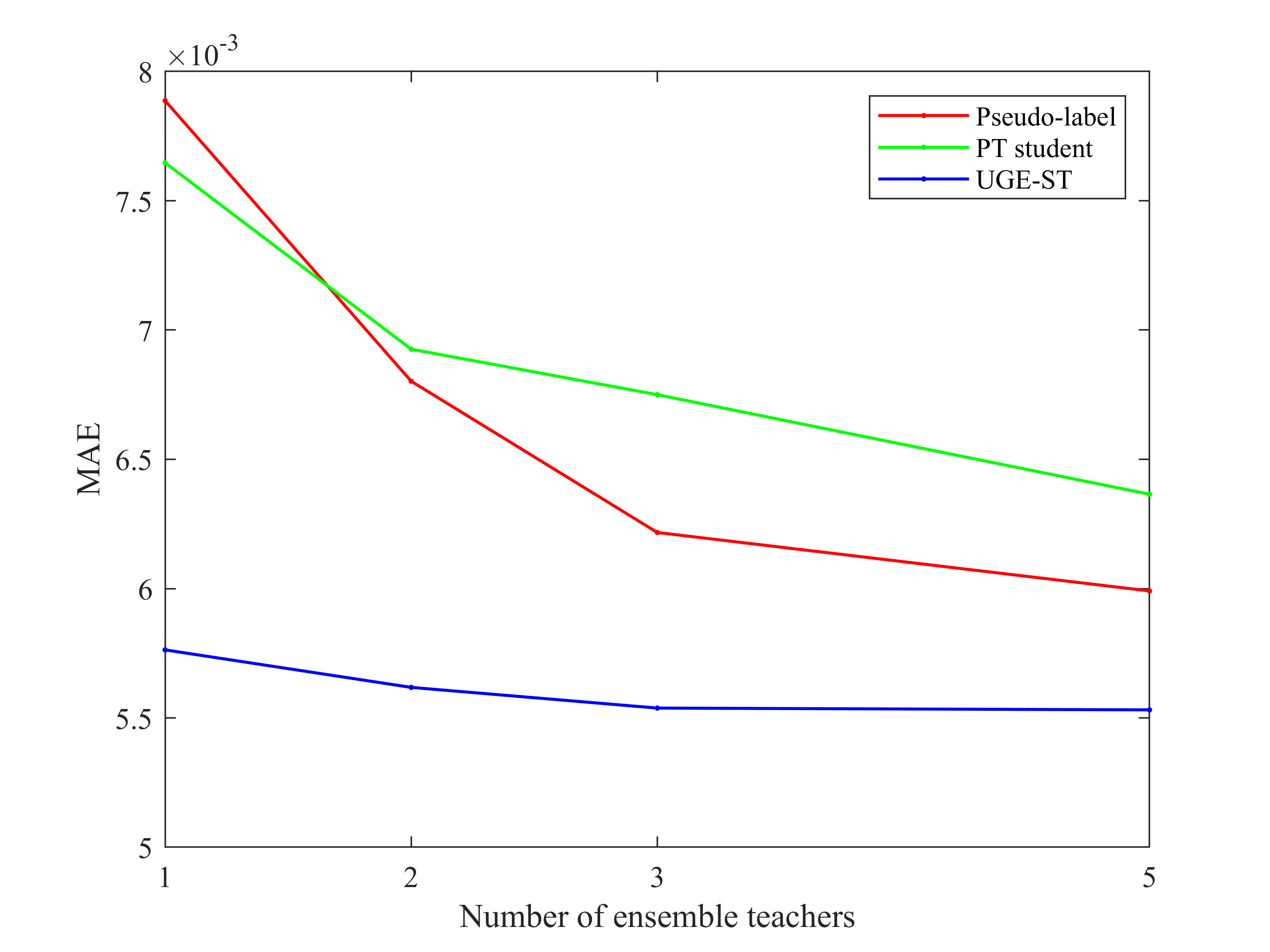} 
	\caption{MAE changes with the number of ensemble teachers}  
	\label{fig:7}
\end{figure}

The effect of the ensemble teachers is shown in Fig.~\ref{fig:7}. We set the number of ensemble teachers to 1 (means no ensemble), 2, 3, and 5, respectively. Pseudo-label means the prediction provided by the ensemble teachers. PT student represent the pre-training student model supervised by the ensemble teachers but without uncertainty guided learning.
UGE-ST is the model trained only with labeled data based on the pre-training student.

The results indicate that with the increase of ensemble teachers number, the accuracy of pseudo-label is significantly improved, further leading to the performance increase of PT student.
Although the UGE-ST is trained based on PT student, the performance is slightly improved. The reason is that this paper uses pseudo-label and labeled data to train the model successively, which reduces the influence of noise on the student model and also reduces the impact of pseudo-label on the final performance of the student model. We finally chose the ensemble number of 3 to balance the accuracy and training cost.

\subsubsection{The influence of pre-training student}
The effect of the pre-training is shown in Tab.~\ref{fig:4}. We set the ensemble number to 1 to avoid the influence of ensemble teachers. Pseudo-label means the prediction provided by the one teacher. PT student represent the pre-training student model supervised by the one teacher but without uncertainty guided learning.
UGE-ST is the model trained only with labeled data based on the PT student.
The results show that the method of pre-training student can well avoid the influence of noise in pseudo-label to the student model. In comparison, the MAE of self-training which combine labeled data and pseudo-label is 6.624e-03, while our method is 5.763e-03.

\setlength{\tabcolsep}{6pt}
\begin{table}[htbp]
	\centering
	\caption{The influence of the pre-training student.}
	\begin{tabular}{c|cccc}
		\toprule
		models & Pseudo-label  & Self-training  & PT student   & UGE-ST \\
		\midrule
		MAE & 7.887e-03 & 6.624e-03 & 7.647e-03 & 5.763e-03 \\
		\bottomrule
	\end{tabular}%
	\label{tab:4}%
\end{table}%
\subsubsection{The influence of uncertainty guided learning}
The influence of uncertainty guided learning is shown in Tab.~\ref{fig:5}.
PT student and UGE-ST achieve better performance when guided by uncertainty. It is worth noting that the gain of PT student from uncertainty guided learning is greater than that of UGE-ST. The pre-training also causes this phenomenon; the impact of the pseudo-label on the final performance of the student model is damped due to the successive training of pseudo-label and labeled data.

\begin{table}[htbp]
	\centering
	\caption{The influence of uncertainty guided learning under different number of ensemle teachers.}
	\begin{tabular}{cc|cccc}
		\toprule
		\multicolumn{2}{c|}{Number of ensemle teachers} & 2    & 3    & 5   & \\
		\midrule
		\multirow{2}{*}{w/o uncertainty
		}    & PT student& 6.926e-03 & 6.750e-03 & 6.365e-03 & \\
		& UGE-ST & 5.641e-03 & 5.538e-03 & 5.531e-03 & \\
		\midrule
		\multirow{2}{*}{w/ uncertainty}   & PT student & \textbf{6.4e-03} & \textbf{6.446e-03} & \textbf{6.243e-03} & \\
		& UGE-ST & \textbf{5.618e-03
		} & \textbf{5.529e-03} & \textbf{5.434e-03} & \\
		\bottomrule
	\end{tabular}%
	\label{tab:5}%
\end{table}%

\section{Conclusion}
In this paper, we propose a semi-supervised method, uncertainty-guided integrated self-training (UGE-ST), which aims to improve the reconstruction performance with few labeled data.
UGE-ST consists of ensemble teachers, uncertainty-guided learning, and pre-trained students.
The ensemble teachers employ ensemble learning to construct multiple teacher models to guide the training of student models jointly and the "collective voting" of the ensemble teachers to mitigate the pseudo-label errors generated by individual teacher models, resulting in the accurate pseudo-label.
Uncertainty-guided learning is based on ensemble teachers to quantify the uncertainty in pseudo-label, forcing students to learn regions with less noise in pseudo-label and avoiding the propagation and accumulation of noise in the student.
Pre-trained student train the student model separately using pseudo-labeled and labeled data, enabling the student model to forget the noise in the pre-learned pseudo-label.
Experiments show that the uncertainty-guided ensemble self-training method proposed in this paper can substantially improve the reconstruction performance of the global physics field with limited labeled data.

\section*{Conflict of interest statement}
On behalf of all authors, the corresponding author states that there is no conflict of interest.

\section*{Replication of results}
The code of the proposed method is publicly available at \href{https://github.com/meitounao110/UGE-ST}{https://github.com/meitounao110/UGE-ST}


\bibliography{mybibfile}

\begin{thebibliography}{10}
\bibitem{tarvainen2017mean}
A.~Tarvainen, H.~Valpola, Mean teachers are better role models: Weight-averaged
  consistency targets improve semi-supervised deep learning results, Advances
  in neural information processing systems 30.
\end{thebibliography}

\end{document}